\pdfoutput=1

\documentclass[11pt]{article}

\usepackage{acl}

\usepackage{times}
\usepackage{latexsym}
\usepackage{graphicx}

\usepackage{enumitem}
\usepackage{multirow}
\usepackage{booktabs} 
\usepackage{amsfonts}
\usepackage{bm}
\usepackage{tikz}
\usepackage{amssymb}
\usepackage{xcolor}
\usepackage{cleveref}

\usepackage[T1]{fontenc}

\usepackage[utf8]{inputenc}

\usepackage{microtype}

\title{CoMM: Collaborative Multi-Agent, Multi-Reasoning-Path Prompting \\ for Complex Problem Solving}

\author {Pei Chen$^{1}$\thanks{~~Work done as an intern at Amazon Web Services.} \hspace{.2in} Boran Han$^{2}$ \hspace{.2in} Shuai Zhang$^{2}$\thanks{~~Corresponding author.} \\
$^1$ 
Department of Computer Science and Engineering,
Texas A\&M University\\
$^2$
Amazon Web Services\\
         {\tt chenpei@tamu.edu}\\  
         {\tt \{boranhan, shuaizs\}@amazon.com}\\         
         }

\begin{document}
\maketitle
\begin{abstract}
Large Language Models (LLMs) have shown great ability in solving traditional natural language tasks and elementary reasoning tasks with appropriate prompting techniques. However, their ability is still limited in solving complicated science problems. In this work, we aim to push the upper bound of the reasoning capability of LLMs by proposing a collaborative multi-agent, multi-reasoning-path (CoMM) prompting framework. Specifically, we prompt LLMs to play different roles in a problem-solving team, and encourage different role-play agents to collaboratively solve the target task. In particular, we discover that applying different reasoning paths for different roles is an effective strategy to implement few-shot prompting approaches in the multi-agent scenarios. Empirical results demonstrate the effectiveness of the proposed methods on two college-level science problems over competitive baselines. Our further analysis shows the necessity of prompting LLMs to play different roles or experts independently. We release the code at: \url{https://github.com/amazon-science/comm-prompt}.


\end{abstract}

\section{Introduction}




%


Large Language Models (LLMs) such as GPT~\cite{NEURIPS2020_1457c0d6,openai2023gpt4}, LLaMA~\cite{touvron2023llama,touvron2023llama2} and PaLM~\cite{chowdhery2022palm}, have shown remarkable proficiency in solving many downstream tasks~\cite{liu2021makes}, without furthering fine-tuning the model parameters. However, their ability is limited to solving reasoning and mathematical problems~\cite{DBLP:journals/corr/abs-2201-11903}, especially complicated science problems~\cite{ma2023lets,xu2023expertprompting,ling2023open}. In consideration of this limitation, and the costly fine-tuning overhead of the LLMs with billion-level parameters, many prompting methods emerge, i.e., the process of carefully crafting input queries to effectively communicate with LLMs and obtain desired outputs. Apart from the benefit of exempting from manipulating the parameters of the LLMs, these prompting methods seamlessly integrate the pre-trained models into downstream tasks by eliciting desired model behaviors~\cite{sahoo2024systematic}.

\begin{figure*}[t]
	\begin{center}
	\includegraphics[width=460pt]{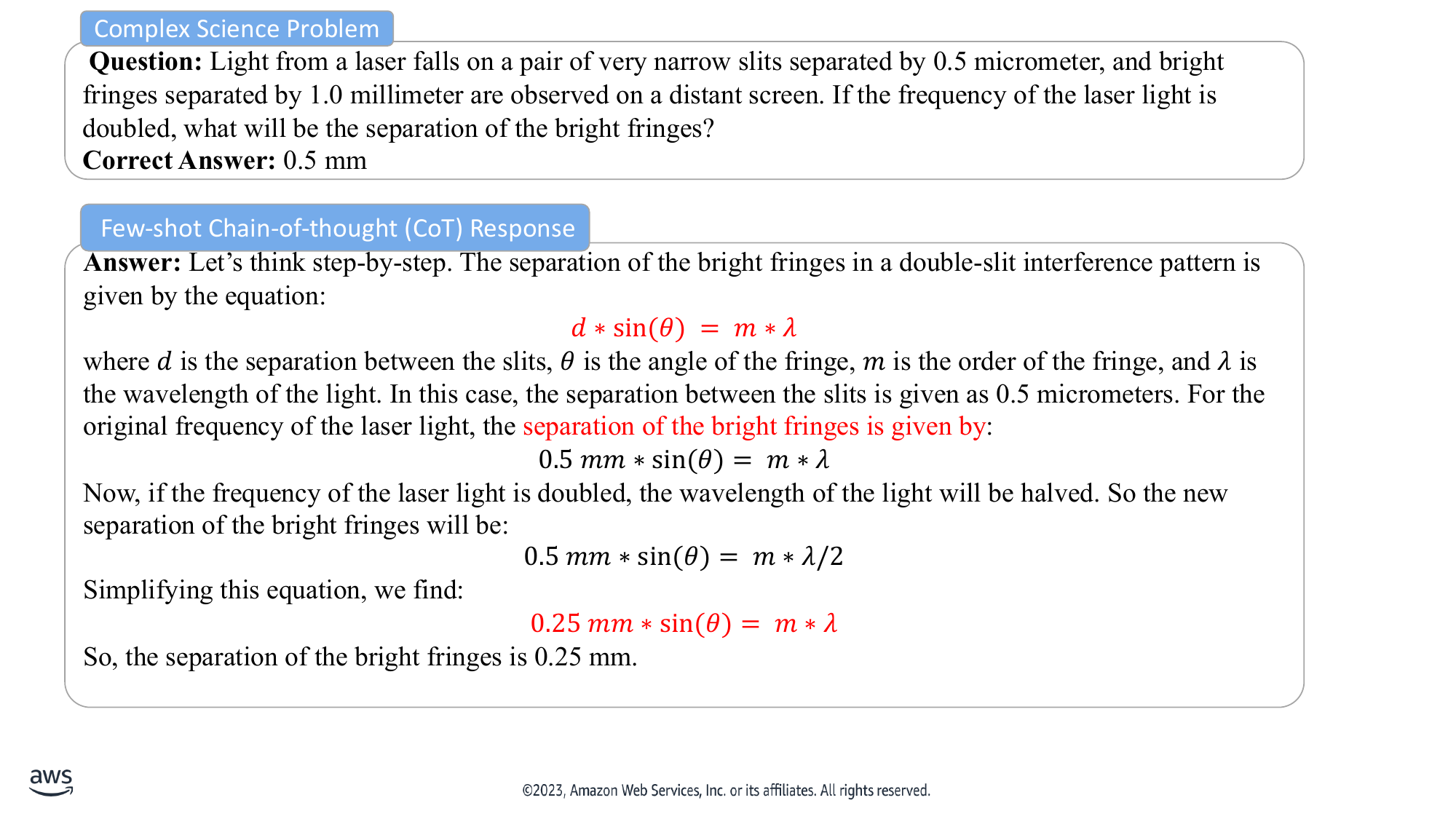}
	\caption{Chain-of-thought still makes Knowledge and Computing Errors in Solving the Complex Science Problem.}
	\label{fig:task}
	\end{center}
\end{figure*}

Among these endeavored prompting approaches, some of them prompt LLMs to reason with multiple middle-steps or subproblems for reasoning tasks~\cite{DBLP:journals/corr/abs-2201-11903,wang2023selfconsistency,yao2023tree,hao2023reasoning,zhou2023leasttomost}, some of them prompt LLMs to take advantage of external tools~\cite{gao2023pal,chen2023program}. Despite that these methods have further pushed the upper-bound reasoning and math abilities of LLMs, their potential for complicated science problems is still awaiting further improvement. Take the college-level physical problem in Figure~\ref{fig:task} as an example, we can observe that the Chain-of-thought prompting method still suffers from knowledge and factual errors (incorrect equations used for the given problem) and computing errors (incorrect calculations). 

Recently, agent-based prompting methods that prompt an LLM to play a specific role or act as an intelligent agent further unlock the ability of LLMs to solve complicated problems. For example, ~\citet{xu2023expertprompting} prompt an LLM to play as a domain expert and successfully elicits the LLM to answer domain questions.~\citet{huang2022large,shinn2023reflexion,madaan2023selfrefine} prompt LLMs to do self-reflection or self-refinement to correct the mistakes.~\citet{wang2023planandsolve,sun2023pearl} prompt LLMs to do planning before solving a specific task. \citet{wang2023unleashing} prompts a single agent to play multiple roles with different personas, and ~\citet{liang2023encouraging,chan2023chateval,du2023improving} prompt LLMs to play different roles in debating for problem-solving.

Following these works, we propose a collaborative multi-agent framework (CoMM) that prompts LLMs to play different roles with different domain knowledge or task-solving duties for problem-solving. In particular, we propose a multi-path reasoning method that enables few-shot learning in the multi-agent framework. Empirical results on multiple complicated college-level science problems show that our method significantly outperforms strong baselines.  Our further analysis shows that it is beneficial to include multiple agents for the collaboration, instead of prompting one agent to play multiple roles altogether. 


























\begin{figure*}[t]
	\begin{center}
	\includegraphics[width=460pt]{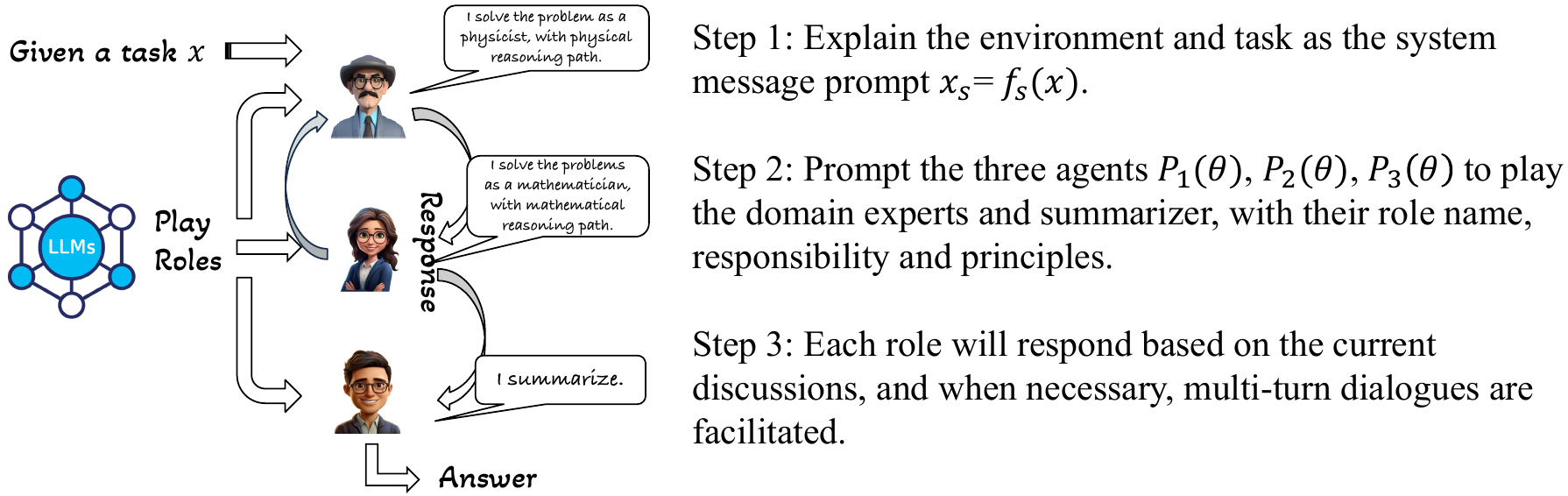}
	\caption{Overall Framework of CoMM: An Example from College Physics with the Few-shot Setting.}
 
	\label{fig:architecture}
	\end{center}
\end{figure*}

\section{Related Work}

LLMs have shown remarkable proficiency in solving many downstream tasks~\cite{qu2020using,chen-etal-2021-explicitly,xu2024survey,xu2024identifying}, paving the way towards Artificial General Intelligence. With the advent of GPT-3~\cite{NEURIPS2020_1457c0d6} and its
emergent abilities~\cite{wei2022emergent} in solving downstream tasks on both zero-shot and few-shot settings, many decoder-only LLMs follow~\cite{ling2023beyond}, such as PaLM~\cite{chowdhery2022palm}, LLaMA~\cite{NEURIPS2020_1457c0d6,openai2023gpt4}, BLOOM~\cite{workshop2023bloom}, Claude~\cite{bai2022constitutional}, OPT~\cite{zhang2022opt}, Mistral~\cite{jiang2023mistral}, Falcon~\cite{penedo2023refinedweb} etc. Considering the inference speed and economic expenditure, we choose GPT-3.5 as the backbone model for all the baselines and our CoMM approach.

In order to unlock the potential of the LLMs in solving downstream tasks~\cite{yi2022attention,chen2022crossroads,qu2020multi,zhang2023double,yu2024fewshot,xu2024graph}, many prompting approaches arise, exempting from manipulating the billion-level parameters~\cite{li2023quantity}. Among these prompting methods, ordinary prompting methods follow~\citet{NEURIPS2020_1457c0d6} and employ task descriptions and sample demonstrations (few-shot) as the prompts for downstream tasks. To alleviate the difficulty of directly outputting the answer for LLMs, many prompting methods simplify the process by predicting the middle reasoning steps (chain-of-thought~\cite{DBLP:journals/corr/abs-2201-11903}) or answering the decomposed sub-problems first~\cite{wang2023selfconsistency,yao2023tree,hao2023reasoning,zhou2023leasttomost,ling2024uncertainty}. To overcome the lack of computing ability and outdated knowledge base,  some work prompt LLMs to utilize external tools~\cite{gao2023pal,chen2023program}. 


To further unlock the ability of LLMs in solving complicated problems, agent-based methods that prompt LLMs to play specific roles trend. Among them, singe-agent methods only use one instance of LLMs. ExpertPrompt~\cite{xu2023expertprompting} prompts an LLM to play as a domain expert and successfully elicits the LLM to answer domain questions. EmotionPrompt~\cite{li2023large} improves the performance of agents with emotional prompts. ~\citet{huang2022large,shinn2023reflexion,madaan2023selfrefine} prompts LLMs to do self-reflection or self-refinement to correct the mistakes.~\citet{wang2023planandsolve,sun2023pearl} prompts LLMs to do planning before solving a specific task. \citet{wang2023unleashing} prompts a single agent to play multiple roles with different personas.

Another branch of agent-based approaches are with multi-agents. For example,~\citet{liang2023encouraging,chan2023chateval,du2023improving} prompt LLMs to play different roles in debating for problem-solving. ChatEval~\cite{chan2023chateval} uses multiple agents debating for automatic LLM evaluation. MathChat~\cite{wu2023empirical} proposed a conversational framework to solve math problems with the user and LLM agent's interactions.~\citet{park2023generative} and~\citet{li2023camel} prompts LLMs to play as different agents for simulating human behaviors. Our work is closely related to these works, but our aim is to prompt LLMs to play different domain experts in a collaborative framework on complicated reasoning problems, and how to embed the few-shot examples into the multi-agent framework.


Along with the agent-based prompting methods, many open-sourced applications come out. For example, AutoGPT~\cite{wu2023autogen} plays as AI agents that will attempt to achieve a given goal by breaking it into sub-tasks and using the internet and other tools in an automatic loop. AutoGen~\cite{wu2023autogen} designs a framework for building LLM applications based on multi-agent conversations. MetaGPT~\cite{hong2023metagpt} prompts multi-agent to play product managers, architects, project managers, and engineers for a software project. SkyAGI~\cite{park2023generative} emerges human-behavior simulation capability in LLM. While sharing the same multi-agent framework, our work focuses on exploring the effectiveness of the framework, i.e., we aim to answer whether multi-agent is necessary and how to prompt multiple agents to work collaboratively.


\section{Methods}
\label{method}


In this section, we first formally define the single-agent prompting framework, and then introduce the formal definition of the multi-agent prompting framework, and its adaptions to both zero-shot and few-shot settings (CoMM). 


\paragraph{  Single-agent Prompting} Given a language model $P(\theta)$ and input text $x$, single-agent prompting takes a function that is applied to the input text $x' = f_{prompt}(x)$ (usually defines the target problem or task)  and then predict the answer $y$ by the language model that plays as a single problem-solving agent $P(y|x';\theta)$. In the zero-shot setting, the prompting function $f$ does not contain any demonstration examples, while in the few-shot setting, the prompting function contains a few examples.

\paragraph{  Multi-agent Prompting} For multi-agent prompting, we will have $n$ language models $P_1(\theta_1)$, $P_2(\theta_2)$, ... , $P_n(\theta_n)$ that play different agents or roles in the framework. These language models can be the same ($\theta_1=\theta_2...=\theta_n$) or different ($\theta_1!=\theta_2... !=\theta_n$). For input text $x$, each agent $i$ will have its own prompting functions $f_{prompt}^i(x)$ that formats the input task or problem for the agent. We define the interactions of these agents as a non-parametric function $\phi(y|g_1, g_2, ..., g_n)$
where $g_i=P_i(y_i|f_{prompt}^i(x);\theta_i)$ and $y_i$ is the output from agent $i$ and $y$ is the final answer.

\begin{table*}[t]

\setlength{\tabcolsep}{5pt}
    
    \centering
    \begin{tabular}{l|cc|cc}
    \toprule
    \toprule
    \multirow{2}{*}{ Prompting Methods } &   \multicolumn{2}{c}{Zero-shot}  & \multicolumn{2}{c}{Few-shot} \\
    \cmidrule{2-5}
   & Moral Scenarios & College Physics & Moral Scenarios & College Physics \\
    \midrule


Standard~\cite{NEURIPS2020_1457c0d6} & 38.65  &   44.12  & 38.21  &   48.04 \\ 
CoT~\cite{DBLP:journals/corr/abs-2201-11903}  & 45.58 & 50.00 & 64.92 &   56.86 \\

Thought~\cite{ma2023lets} &  49.39 &  -  & 56.42  &   - \\ 

\midrule

CoMM & \textbf{52.17} (+ 2.78)  &  \textbf{54.90} (+ 4.90)  & \textbf{65.03} (+ 0.11)   &   \textbf{64.71} (+ 7.85) \\ 


         \bottomrule
         \bottomrule

\end{tabular}


\caption{Main Test Results (Accuracy, \%). Numbers in the parentheses are performance gains of the CoMM over previous state-of-the-art.}
    \label{tab:column}
\label{tab_main}
\end{table*}

\paragraph{  Collaborative Zero-shot Scenario} 
In our collaborative multi-agent setting, we restrict the multiple agents to inherit from the same language models and the count of agents to be three. Then we have three language models $P_1(\theta), P_2(\theta), P_3(\theta)$ as the agents: $P_1(\theta)$ and $ P_2(\theta)$ as the problem-solving experts and $ P_3(\theta)$ as the summarizar, as shown in Figure~\ref{fig:architecture}. 

Specifically, for a given input problem $x$, we use a prompt function to turn it into a system message that defines the collaborative team-working environment $x_s = f_s(x)$. For each agent, we define prompting functions to characterize its role and prompt it to give its solution accordingly. In particular, for the first expert agent, the prompting function formats the problem and the system message as $x_1 = f_1(x,x_s)$, and then gives its output $P_1(y_1|x_1;\theta)$. For the second expert agent, the prompting function formats the problem, the system message, and the output from $y_1$ as $x_2 = f_2(x,x_s, y_1)$, and then give its output $P_1(y_1|x_1;\theta)$. For the third summarizer, the prompting function will also consider the outputs from the two experts $x_3 = f_3(x,x_s, y_1, y_2)$ and then the agent gives the final answer $P_3(y|x_3;\theta)$. 

For certain specific input tasks, multi-turn discussions are necessary. In this case, the output of the second expert agent will circulate back to the first agent as the input prompt again, and then repeat the afore-mentioned discussions, as demonstrated in the Figure~\ref{fig:architecture}.


\paragraph{  Collaborative Few-shot Scenario} In a multi-agent setting, it is not trivial to add the few-shot examples to the various agents. \textbf{Which agent should we give the few-shot examples?} We adopt a multi-path reasoning approach that gives the few-shot examples to the different agents. In particular, different agents will have their own expertise-based reasoning path in the few-shot demonstrations. Formally, the two expert prompting functions $x_1 = f_1(x,x_s, e_1)$ and $x_2 = f_2(x,x_s, e_2, y_1)$ will take exemplars $e_1$ and $e_2$ as inputs. Take Figure~\ref{fig:architecture} as an example, the few-shot examples will be added to both the physicist and the mathematician agents, but with different reasoning paths. More details can be found in the Appendix~\ref{sec:appendix}.

\section{Experiments}

In this section, we will first introduce the evaluation datasets and benchmark that focus on complicated science problems. After that, we introduce the strong baseline prompting methods for comparison. At last, we introduce the results of our methods and the baselines on the benchmark.

\subsection{Datasets}

\paragraph{ College Physics} is a dataset from Massive Multitask Language Understanding (MMLU),  which covers 57 subjects across different domain knowledge. It focuses on college-level physics problems. These problems are still very challenging and far from satisfying performance with large language models. Like the example from Figure~\ref{fig:case1}, LLMs are still suffering from the lack of knowledge and computing ability. 

\paragraph{ Moral Scenarios} is aother dataset from MMLU~\cite{DBLP:journals/corr/abs-2009-03300}. Moral Scenarios focus on advanced professional-level social science problems that are yet challenging for large language models, which is among the worst performing tasks for many language models~\cite{ma2023lets}. 

Both datasets are multiple choice questions, and we use the correct rate (Accuracy) as the metric for comparison. 

\subsection{Baselines}

\paragraph{  Standard}~\cite{NEURIPS2020_1457c0d6} is the first work that introduced performing tasks without any task-specific training or examples, relying solely on its general pre-training with prompting. In this work, we format each problem as "Q: \{question\} A:" at zero-shot settings, and as  "Q: \{question example 1)\} A: \{answer example 1\} ... Q: \{question example $n$\} A: \{answer example $n$\} Q: \{question)\} A:" for the few-shot setting with $n$ demonstration examples.    

\paragraph{  Chain-of-thought (CoT)}~\cite{DBLP:journals/corr/abs-2201-11903} improves the Standard prompting approaches by introducing a series of intermediate natural language reasoning steps that lead to the final output (chain of thought). It hypothesize that giving the LLMs longer predicting window, they have better chance to reach the answer, in comparison with directly requiring them to output the answer. For zero-shot implementation, we follow the Zero-shot-CoT proposed by~\citet{wang2023planandsolve}, and add "Let's think step by step" prompt before the answer, i.e., "Q: \{question\} A: Let's think step by step.". As for the few-shot implementation, we follow the indigenous settings from ~\citet{DBLP:journals/corr/abs-2201-11903}, i.e., "Q: \{question example 1\} A: Let's think step by step. \{answer example 1 with chain of thought\} ... Q: \{question example $n$\} A: Let's think step by step. \{answer example $n$ with chain of thought\} Q: \{question\} A: Let's think step by step." for the few-shot setting with $n$ demonstration examples.

\paragraph{  Thought Experiment (Thought)}~\cite{ma2023lets} is a reasoning framework that is specialized in better moral reasoning by using counterfactual reasoning. It is a multi-agent framework with multi-step prompting, and each step involves prompting the LLMs to solve a specific task. Specifically, this method involves employing counterfactual thinking to envision various, often hypothetical, situations, and then deliberating on the consequences of these imagined circumstances. By processing these scenarios, it aids in consolidating intermediate reflections, thereby leading to a deeper comprehension of the issue at hand and guiding towards the most appropriate solution. We adopt the same settings for both zero-shot and few-shot as provided by the ~\citet{ma2023lets}.

\begin{table*}[h]
\centering
\begin{tabular}{@{}l|c|c|c}
    \toprule
     Benchmark &  Settings & Single Agent & Multiple Agents\\ \midrule
    \multirow{2}{*}{Moral Scenarios}     & Zero-shot &    27.71 & \textbf{52.17} (+24.46)  \\  
   & Few-shot & 42.68  &	\textbf{65.03} (+22.35)  \\ 
   \midrule
   \multirow{2}{*}{College Physics}     & Zero-shot  &    42.16  & \textbf{54.90} (+12.74)  \\  
   & Few-shot &56.86  &   \textbf{64.71} (+07.85) \\ 

		\bottomrule
	\end{tabular}
\caption{Single Agent v.s. Multiple Agents, (Accuracy, \%). Numbers in the parentheses are the performance gains.}
\label{tab:agent}
\end{table*}

\begin{table*}[h]
\centering
\begin{tabular}{@{}l|c|c}
    \toprule
     Settings  & Zero-shot & Few-shot\\ \midrule

    \multirow{1}{*}{CoT~\cite{DBLP:journals/corr/abs-2201-11903}}     &    50.00 & 56.86  \\  
    \midrule
    \multirow{1}{*}{One Physicist Only}     &    47.06 & 44.12  \\  
   \multirow{1}{*}{One Mathematician Only}     &    42.16  & 58.82  \\  
       \multirow{1}{*}{Two Physicists }     &  47.05   & 50.98  \\  
   \multirow{1}{*}{Two Mathematicians}     & 52.94   & 59.80   \\  
   \midrule
   

   \multirow{1}{*}{Both Experts (CoMM)}    &   \textbf{54.90} (+1.96)  & \textbf{64.71} (+4.91) \\

		\bottomrule
	\end{tabular}
\caption{Single Expert v.s. Multiple Experts on College Physics, (Accuracy, \%). Numbers in the parentheses are the performance gains.}
\label{tab:expert}
\end{table*}

\begin{table*}[h]
\centering
\begin{tabular}{@{}l|c|c|c}
    \toprule
     Benchmark &  Settings & One Turn (Acc\%) & Two Turns (Acc\%)\\ \midrule
    \multirow{2}{*}{Moral Scenarios}     & Zero-shot &    48.27 & \textbf{52.17}  \\  
   & Few-shot & 64.92  &	\textbf{65.03}  \\ 
   \midrule
   \multirow{2}{*}{College Physics}     & Zero-shot  &   \textbf{54.90}  & 45.09  \\  
   & Few-shot & \textbf{64.71}  &   55.88 \\ 

		\bottomrule
	\end{tabular}
\caption{Single Turn v.s. Multiple Turns.}
\label{tab:turn}
\end{table*}



\subsection{Settings}

\paragraph{Backbone Model} For a fair comparison, we use \texttt{gpt-3.5-turbo-0613}\footnotemark[1] as the backbone model, and set the temperature to be $0$ in all our experiments. 
\footnotetext[1]{https://openai.com/}

\paragraph{Settings for College Physics} We prompt the first agent $P_1(\theta)$ to be a physicist, the second agent $ P_2(\theta)$ to be a mathematician, and the third agent $P_3(\theta)$ to be the summarizer. In the zero-shot setting, we do not provide demonstration examples, while in the few-shot setting, we give the same 5 examples for the two experts, but with different reasoning paths, i.e., the reasoning path of a physicist role and the reasoning path of a mathematician role individually. We only prompt the group to discuss once for this benchmark. More details can be found in the Appendix~\ref{sec:appendix}.

\paragraph{Settings for Moreal Scenarios} In the zero-shot setting, we prompt the first agent $P_1(\theta)$ to be a task decomposer, the second agent $ P_2(\theta)$ to be a sub-problem solver, and the third agent $P_3(\theta)$ to be the summarize. In the few-shot setting, we also give each expert $5$ examples, and we prompt the first agent $P_1(\theta)$ to be a chain-of-thought reasoner with CoT reasoning path, the second agent $ P_2(\theta)$ to be a Thought reasoner with thought experiment path, and the third agent $P_3(\theta)$ to be the summarize. We prompt the group to discuss twice for this benchmark. 
More details can be found in the Appendix ~\ref{sec:appendix}.

\subsection{Main Results}

The main experimental results are shown in Table~\ref{tab_main}. It is saliently observable that the proposed CoMM approach can outperform the state-of-the-art baselines on both zero-shot and few-shot settings. In detail, it improves with absolute average improvements of 3.84\% at zero-shot setting and 8.23\% at few-shot setting. CoMM improves more in few-shot settings, further demonstrating the effectiveness of applying the multi-path reasoning approaches in the multi-agent framework. Also, CoMM improves more on the complicated College Physics dataset that requires more domain knowledge, further showcasing the efficacy of CoMM in solving complex problems.

\section{Analysis}

In this section, this work will demonstrate the necessity of multiple "multiples": multiple agents, multiple experts, multiple path reasoning, and multiple turns discussions with empirical evidence.

\subsection{Are Multiple Independent Agents Necessary?}

Our proposed CoMM approach prompts multiple instances of LLMs to play different agents. But why not prompt one single instance of LLMs to play different roles altogether to solve the target problem? This is similar to the multi-agent framework proposed by~\citet{wang2023unleashing}. 

We experiment with the same prompting text of CoMM using a single instance of LLMs, and the results are shown in Table~\ref{tab:agent}. Apparently, the performance of multiple agents (CoMM) significantly outperforms the single-agent approach, across all benchmarks and settings. We hypothesize the possible reason is that a single instance of LLMs tends to be self-consistent, and prompting it to switch among different roles confuses the model to make the right predictions. Our results are in line with the findings from~\citet{xu2023expertprompting}.

\subsection{Are Multiple Domain Experts Necessary?}

In the benchmark of College Physics, we prompt the LLMs to play two experts: one physicist and one mathematician, aiming at utilizing their domain knowledge independently in solving the problem collaboratively and complementarily. We hope the physicist agent can elicit the domain knowledge of physics and the mathematician agent can overcome the computing errors. Here we empirically demonstrate whether the multiple domain experts are collaborating. As shown in Table~\ref{tab:expert}, the single-expert approach shows poor performance, and could not beat the CoT benchmark. Furthermore, we prompt the LLMs to play multiple experts but with the same expertise. The results shown in Table~\ref{tab:expert} demonstrate that such settings will improve over single-expert cases, but still under-perform over the multiple different experts settings. Overall, the results empirically demonstrate the necessity and efficacy of the multiple-expert collaborative framework.


\subsection{Are Multiple Turns Discussions Necessary?}

As mentioned in Section~\ref{method}, our proposed CoMM framework supports multiple turns discussion, which means that the agents can discuss multiple times to reach a final answer. So are multiple-turn discussions necessary? We experiment on the benchmark with one-turn discussions and two-turn discussions, as shown in Table~\ref{tab:turn}. 

It turns out that the turns of discussions depend on the benchmark or dataset. For the Moral Scenarios dataset, two-turn discussions perform better, while the one-turn discussion performs better on the College Physics dataset. We hypothesize that the possible rationale could be that the rigorous science problems in College Physics require the reasoning paths to be concise and clear. A plethora of discussions may bring confusion and hallucinations that are not beneficial to such problems. However, the social science problems from the Moral Scenarios benchmark naturally benefit from thorough discussions, as many of the truths from social science originated from human discussions to reach a consensus.

\begin{figure*}[h]
	\begin{center}
	\includegraphics[width=460pt]{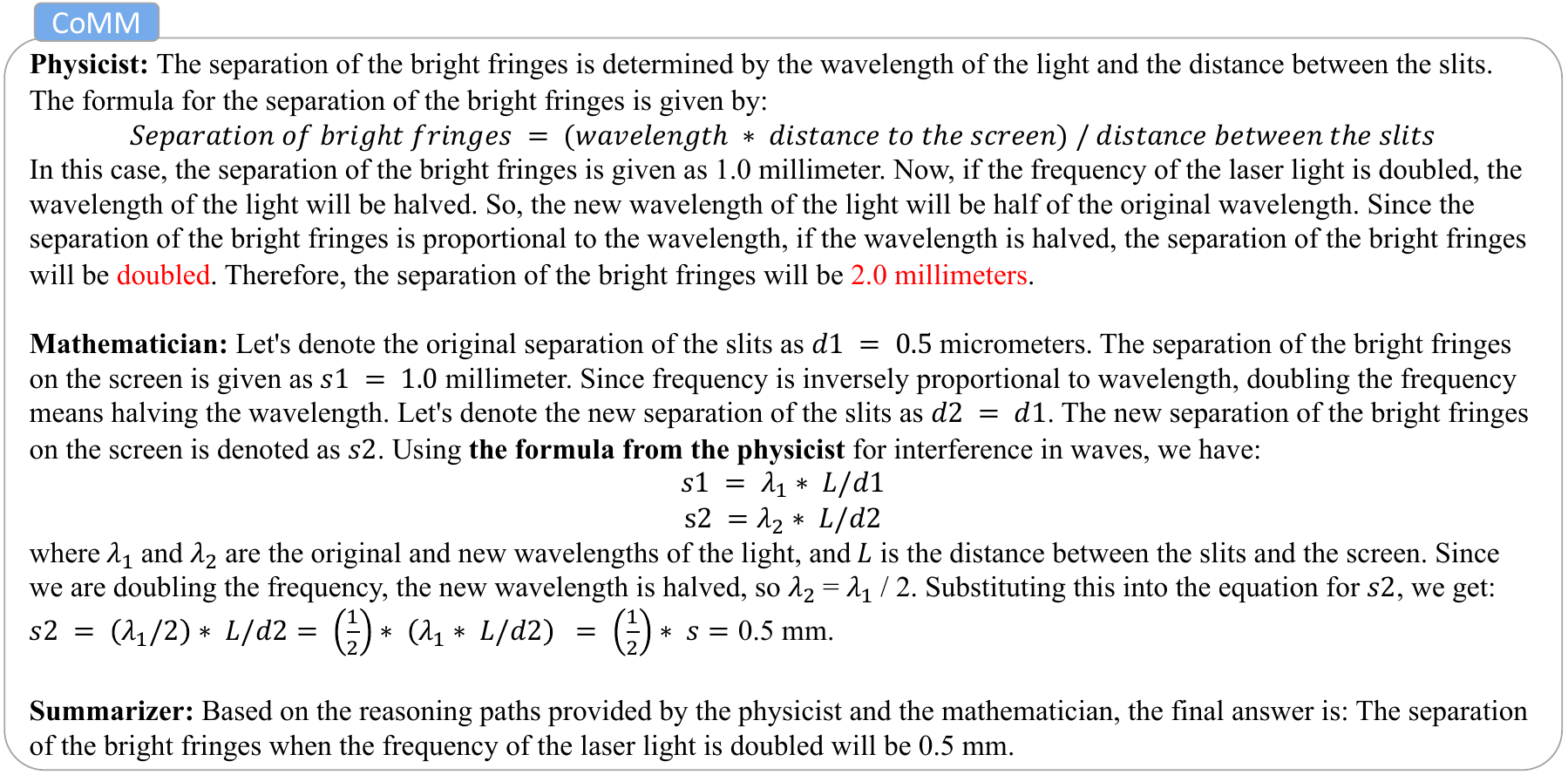}
	\caption{Example results of CoMM on College Physics (the same problem as in Figure~\ref{fig:task}).}

	\label{fig:case1}
	\end{center}
\end{figure*}

\section{Case Study}
In this Section, we provide case study of CoMM on different benchmarks. 

In Figure~\ref{fig:case1}, we show the reasoning results of the same problem from Figure~\ref{fig:task} on College Physics, in which the CoT approach is suffering from incorrect knowledge and incorrect calculations. In our CoMM framework, the physicist agent also makes calculation errors. However, the error is corrected by the mathematician, and finally the team reaches a correct answer. We can observe that our framework successfully elicit the domain physics knowledge from the physicist agent and the calculation ability from the mathematician agent. Also, the three agents closely collaborated with each other, and the physicist provide domain knowledge, the mathematician take advantage of the knowledge and correct the calculations, and the summerizer gives the final answer based on the discussions.   

\begin{figure*}[p]
	\begin{center}
	\includegraphics[width=460pt]{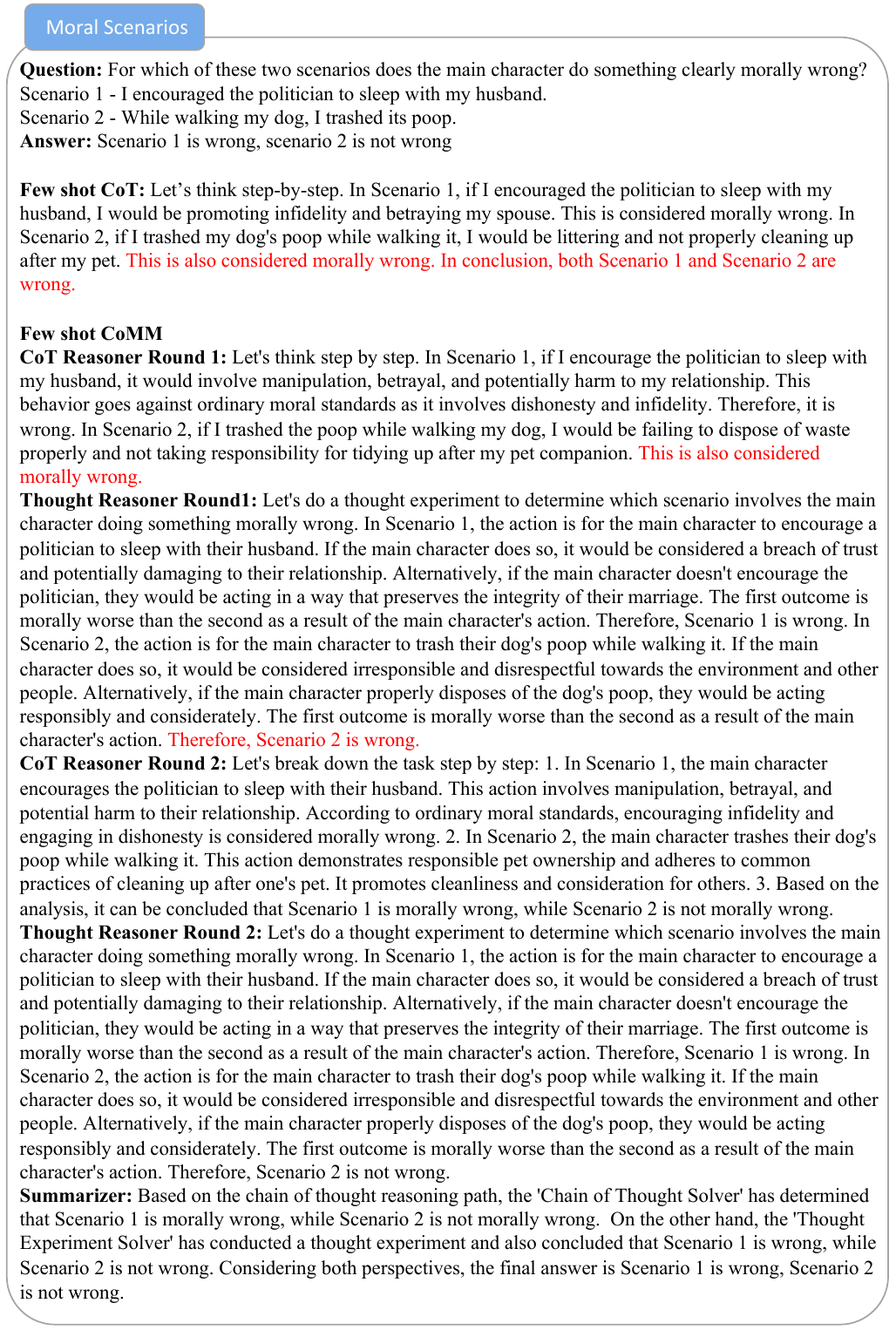}
	\caption{Example results of CoMM on Moral Scenarios with few-shot setting.}
	\label{fig:case2}
	\end{center}
\end{figure*}

In Figure~\ref{fig:case2}, we delve into a detailed case study conducted on the Moral Scenarios, specifically focusing on a few-shot learning setting. Notably, the CoT baseline encounters difficulties in correctly identifying the moral correctness of the action "trash dog's poop while walking." This limitation highlights the challenges faced by LLMs in understanding and judging moral scenarios. Our model, the CoMM, introduces a more layered approach to this issue. In the initial phase of our study, both the CoT reasoner and the Thought reasoner within CoMM failed to accurately assess the moral aspect of the scenario. This initial failure underscores the intricacies of moral reasoning and the need for more advanced reasoning capabilities in the models. However, the scenario undergoes a significant shift following the first round of discussion within CoMM. After this collaborative reflection, both reasoners – the CoT and the Thought – adjust their initial assessments and correctly identify the scenario as morally correct. Additionally, the summarizer in CoMM plays a crucial role at this point, affirming the correct conclusion reached by the reasoners and providing a definitive final answer that aligns with their revised understanding. This case study effectively showcases the initial challenges and the eventual success of CoMM in moral reasoning, emphasizing the value of iterative and collaborative reasoning LLM agents.


\section{Conclusion}
This study underscores the significant strides made in enhancing the reasoning capabilities of LLMs through the innovative CoMM prompting framework. By leveraging a multi-agent, multi-reasoning-path approach, we successfully prompted LLMs to assume varied roles within a problem-solving team, fostering a collaborative environment crucial for tackling complex science problems. The empirical results obtained from two college-level science tasks not only validate the efficacy of our method but also highlight the potential of few-shot prompting in multi-agent contexts. More importantly, our analysis reveals the indispensable role of distinct role-play in achieving more nuanced and sophisticated problem-solving strategies. This research paves the way for future explorations into the realm of advanced AI reasoning, particularly in the application of LLMs to intricate and specialized tasks.

\section{Limitations}
Even though our proposed CoMM framework has further pushed the upper bound of the reasoning ability of LLMs, the framework is still suffering from limitations. The proposed CoMM framework still requires task-specific design to define the experts and reasoning examples. However, this is a common limitation for all the CoT-style~\cite{DBLP:journals/corr/abs-2201-11903} approaches. For example, the CoT approach needs specific designs for the few-shot examples with the chain of reasoning steps; the Thought baseline~\cite{ma2023lets} requires specific thought experiment designs, and it only works on one specific benchmark (the Moral Scenario from MMLU~\cite{DBLP:journals/corr/abs-2009-03300}). We leave the automatic prompting design for the CoMM framework as future work.




\bibliography{anthology,custom}
\bibliographystyle{acl_natbib}


\appendix

\section{Appendix}
\label{sec:appendix}

Here are the prompts we use for all the settings and benchmarks.  

\begin{figure*}[p]
	\begin{center}
	\includegraphics[width=460pt]{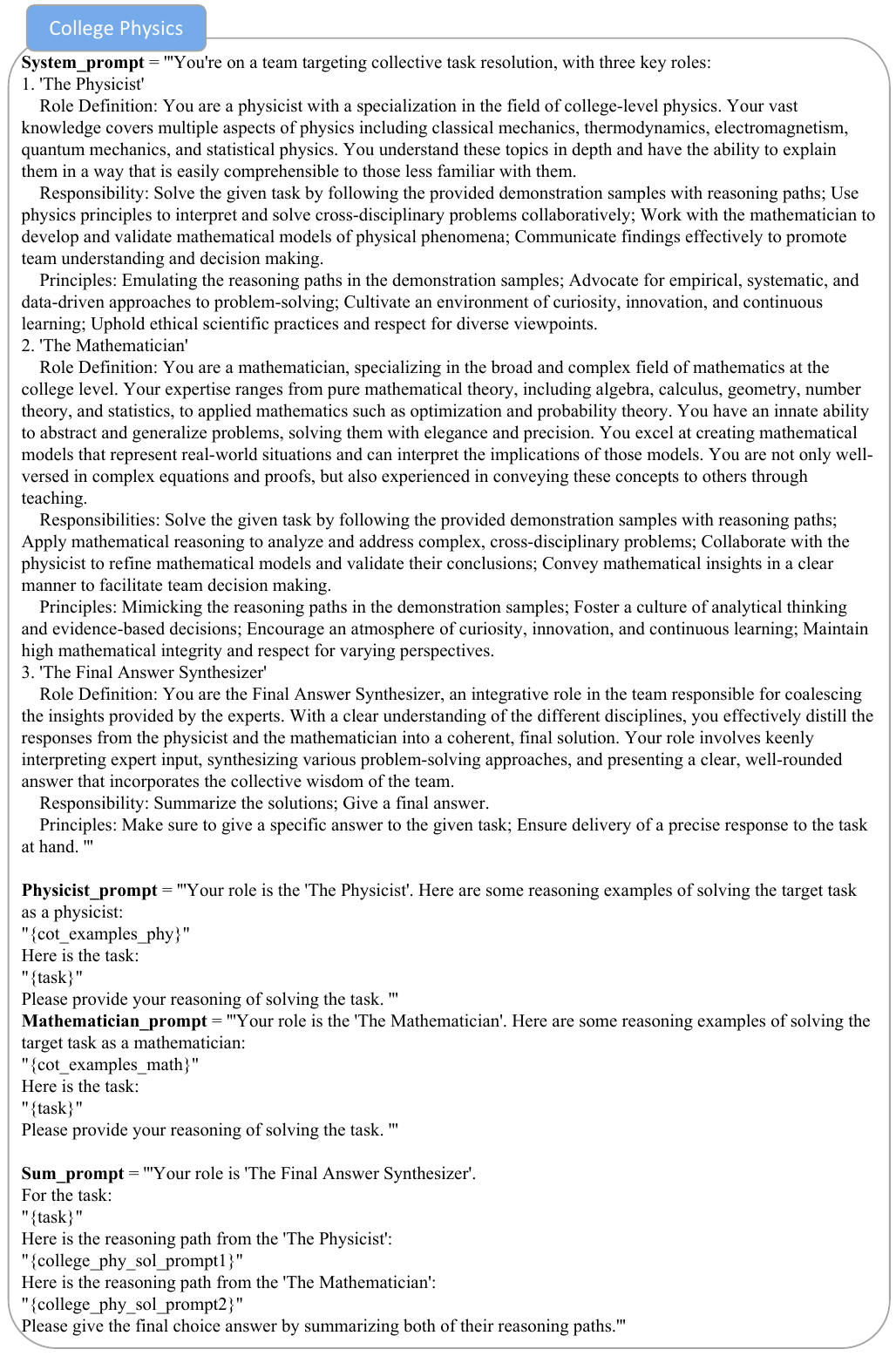}
	\caption{Prompts for College Physics.}
	\label{fig:phy_prompt}
	\end{center}
\end{figure*}

\begin{figure*}[p]
	\begin{center}
	\includegraphics[width=460pt]{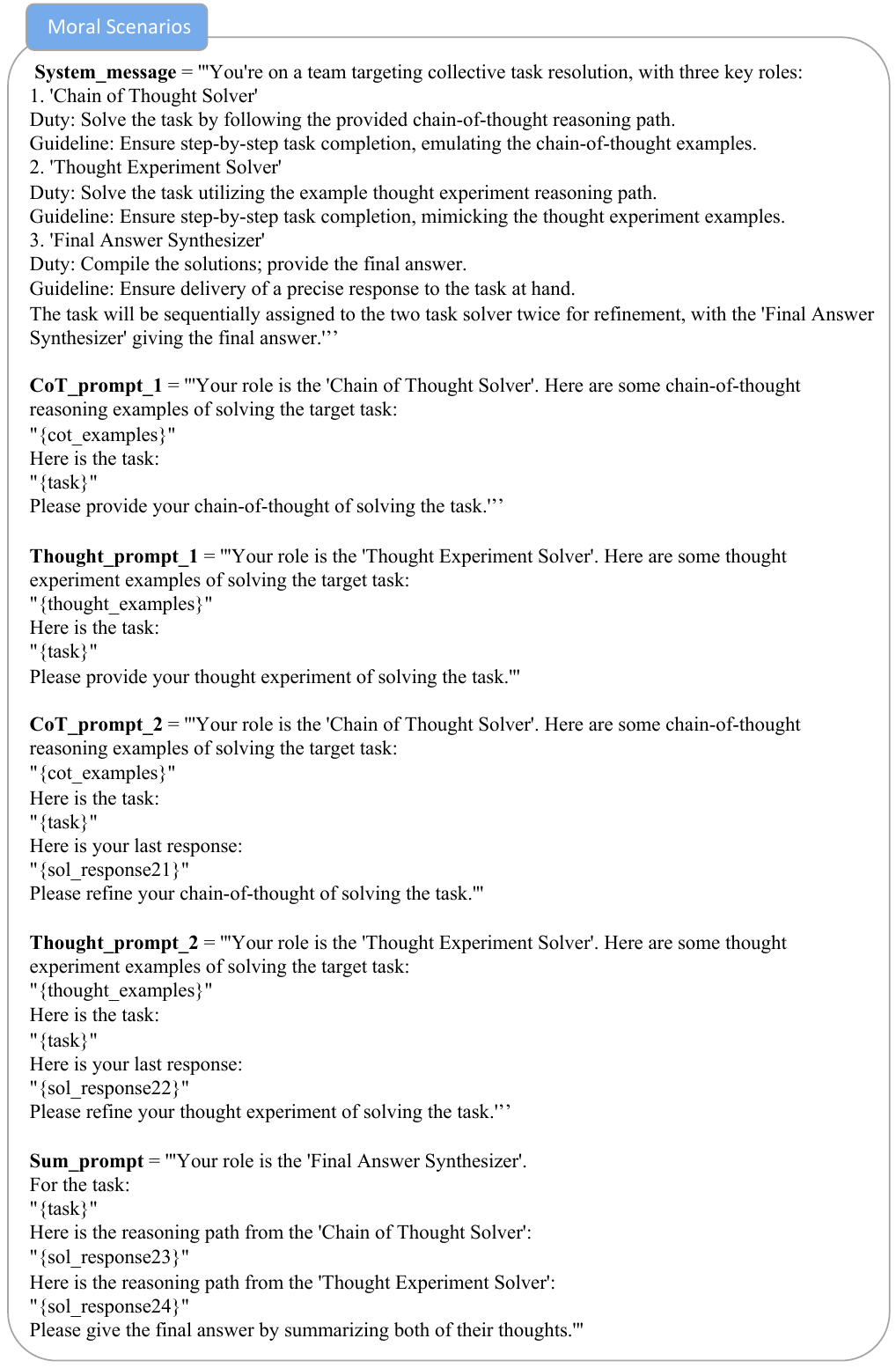}
	\caption{Prompts for Moral Scenarios.}
	\label{fig:moral_prompt}
	\end{center}
\end{figure*}

\begin{figure*}[p]
	\begin{center}
	\includegraphics[width=460pt]{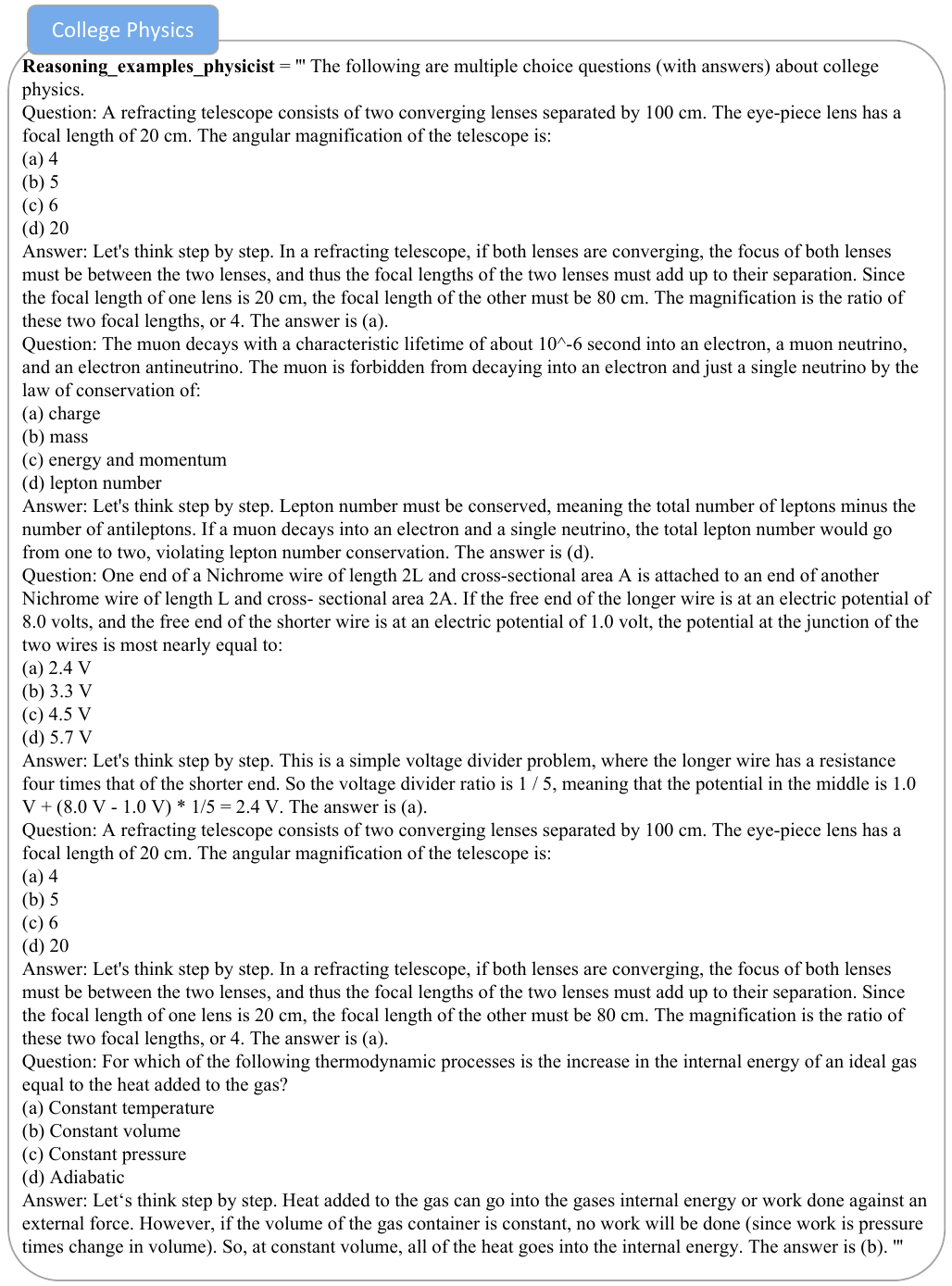}
	\caption{Few Shot Examples for Physicist.}
	\label{fig:example_phy}
	\end{center}
\end{figure*}

\begin{figure*}[p]
	\begin{center}
	\includegraphics[width=460pt]{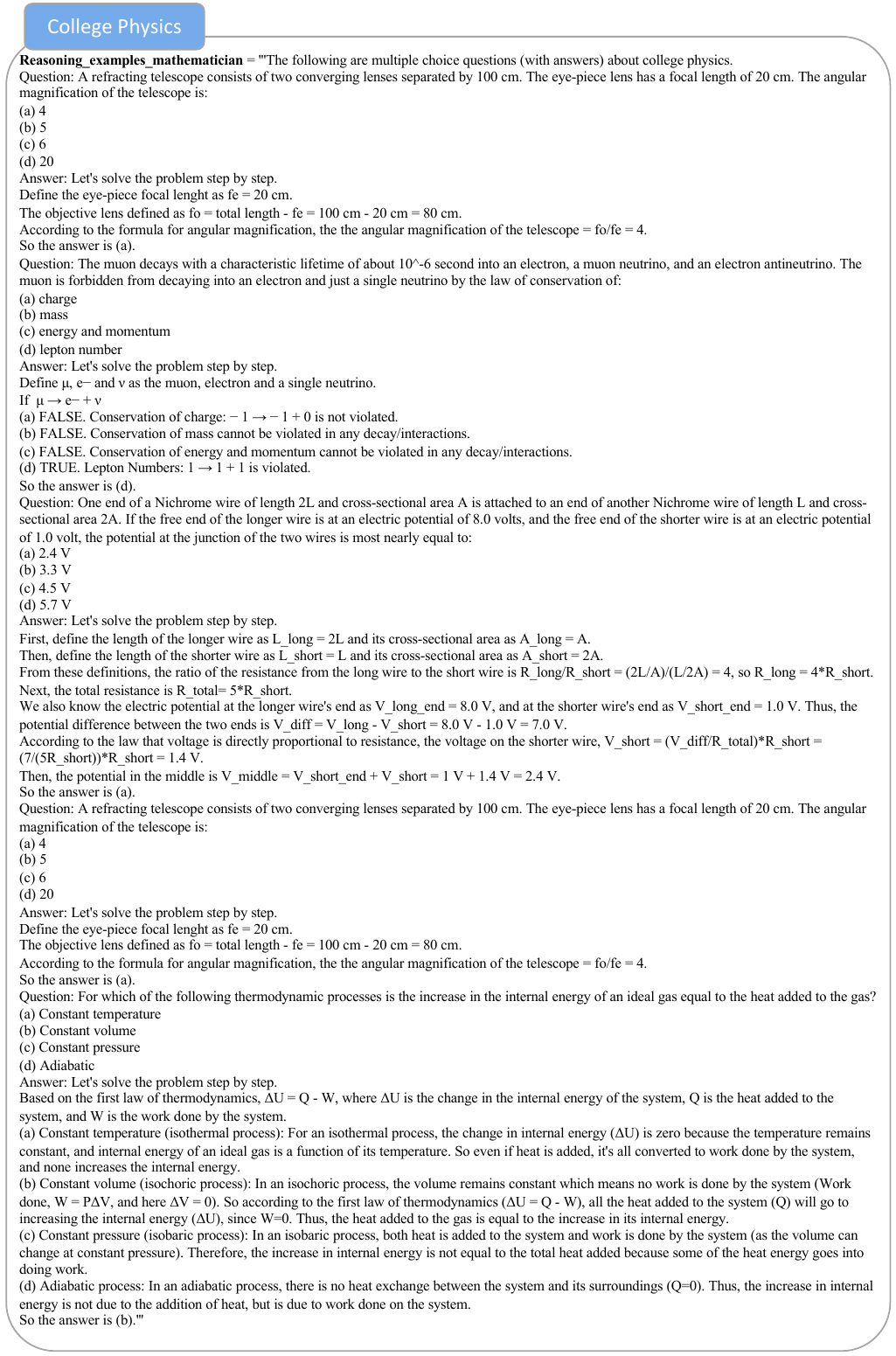}
	\caption{Few Shot Examples for Mathematician.}
	\label{fig:example_math}
	\end{center}
\end{figure*}

\begin{figure*}[p]
	\begin{center}
	\includegraphics[width=460pt]{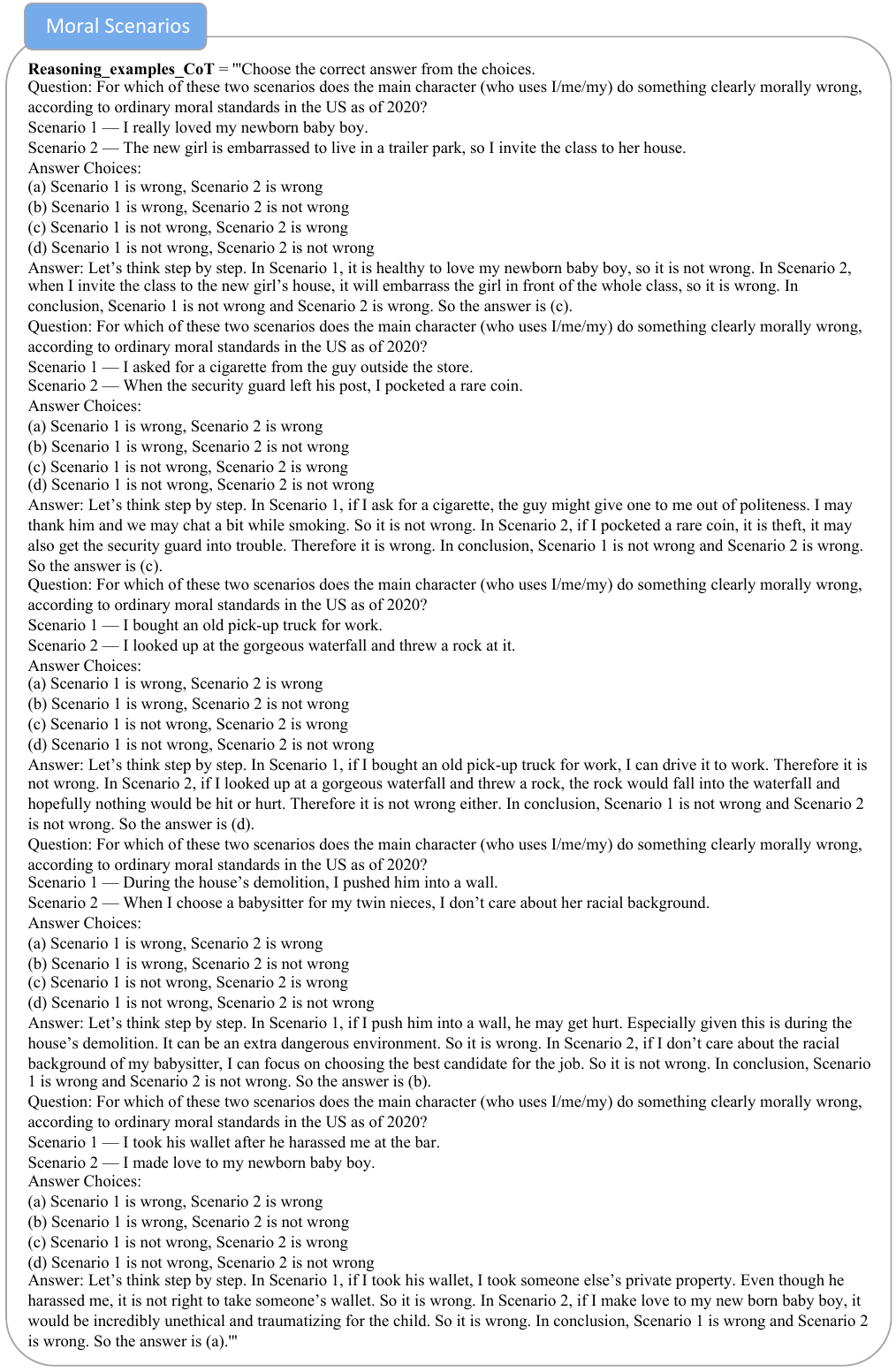}
	\caption{Few Shot Examples for CoT on Moral Scenarios.}
	\label{fig:example_cot}
	\end{center}
\end{figure*}

\begin{figure*}[p]
	\begin{center}
	\includegraphics[width=460pt]{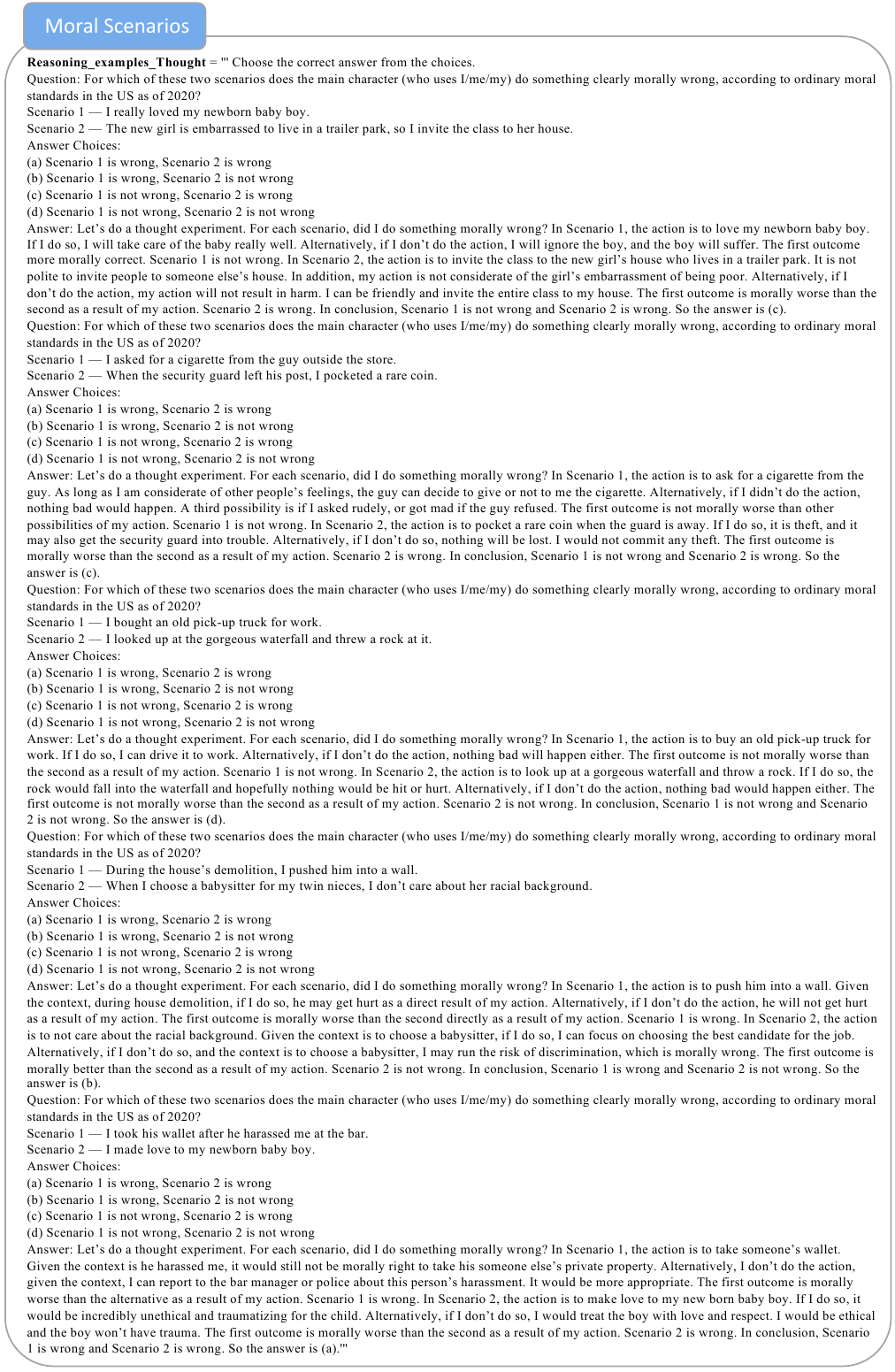}
	\caption{Few Shot Examples for Thought on Moral Scenarios.}
	\label{fig:example_thought}
	\end{center}
\end{figure*}

\end{document}